\documentclass[runningheads]{llncs}
\usepackage[T1]{fontenc}
\usepackage{graphicx}
\usepackage{mathtools}
\usepackage{amsmath}
\usepackage{svg}
\usepackage[dvipsnames]{xcolor}
\usepackage{ulem}
\usepackage{float}
\usepackage{wrapfig}
\usepackage{booktabs}
\usepackage{multirow}
\usepackage{cmap}
\usepackage{orcidlink}

\begin{document}
\title{Beyond Reprojection Error: Camera Calibration with 3D Targets}
%
%
\author{
	Dennis Ruppel\inst{1}\orcidlink{0009-0008-4348-9308}
	\and Hasan Kutlu\inst{1,2}\orcidlink{0000-0003-4969-9564}
	\and Kai A. Neumann\inst{1}\orcidlink{0000-0002-4359-5329}
	\and Martin Knuth\inst{1}
	\and Pedro Santos\inst{1}\orcidlink{0000-0003-1813-1714}
	\and Andreas Weinmann\inst{2}\orcidlink{0000-0002-4969-7609}
	\and Arjan Kuijper\inst{1,3}\orcidlink{0000-0002-6413-0061}
}

\authorrunning{D. Ruppel, H. Kutlu et al.}
%
\institute{
	Fraunhofer Institute for Computer Graphics Research, Darmstadt
	\and Computer Vision, Imaging and Data Analysis Group, Technical University of Applied Sciences Würzburg-Schweinfurt
	\and Mathematical and Applied Visual Computing, Technical University Darmstadt}
\maketitle

\begin{abstract}
\vspace{-1.4cm}
\noindent
In 3D reconstruction, camera calibration is an essential element for achieving high fidelity and accuracy of the reconstructed geometry. While existing approaches rely upon 2D planar calibration, this work proposes a framework tailored for 3D reconstruction that is based on predicting scene rays, which adds flexibility to the reconstruction pipeline and enables the use of recent advances in camera models. Novel metrics, reconstruction and intersection error, derived from predicted scene rays are employed in combination with a bootstrapping procedure that statistically evaluates different calibration objects and calibration pipelines for both intrinsic and extrinsic camera parameters. The results show that the generalized distortion model more faithfully captures physical camera effects and yields an improvement in calibration accuracy. Reprojection error is shown to be a potentially misleading indicator of 3D accuracy, and the proposed ray-based metrics provide a more holistic assessment. An icosahedron calibration target is designed to enrich calibration information for 3D reconstruction together with a ring-feature-based detector. The icosahedral target yields approximately 40\,\% lower mean intersection and more stable calibration across bootstrap trials on synthetic data, while real-data performance demands very tight fabrication tolerances.
\vspace*{-0.3cm}
\begin{figure*}[!b]
	\centering
	\includegraphics[width=0.9\textwidth]{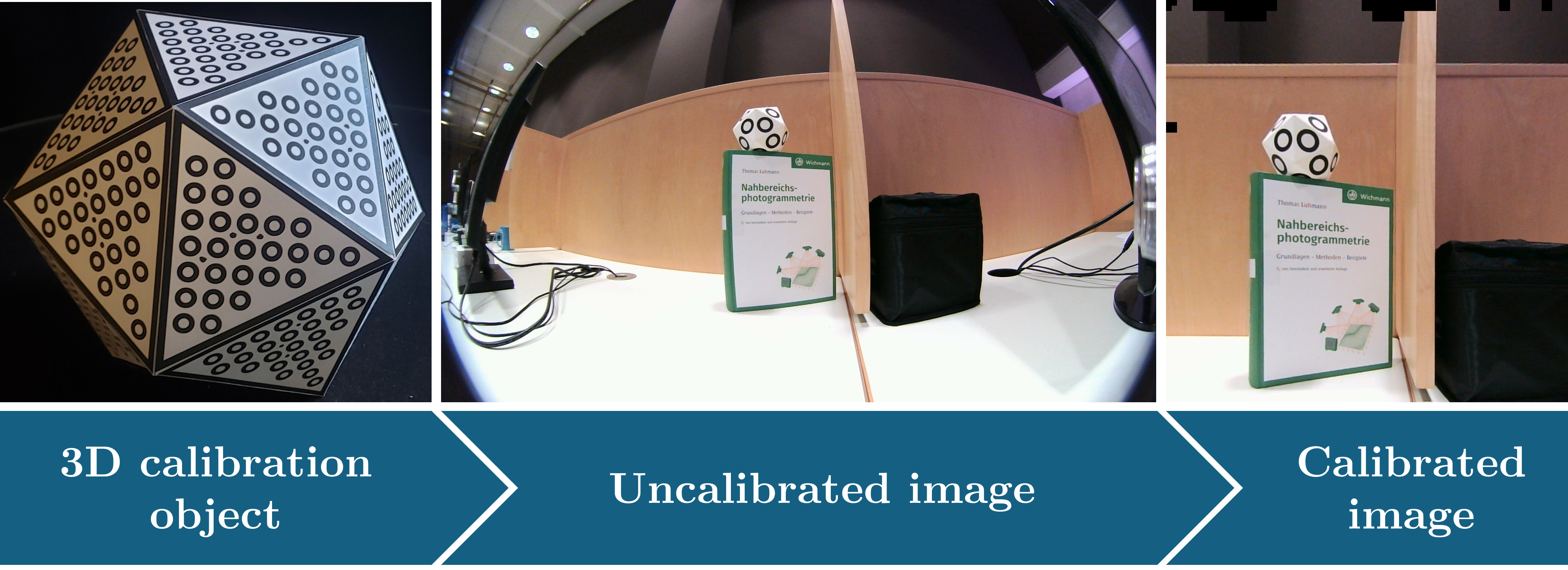}
	\caption{Proposed 3D calibration object: uncalibrated vs.\ calibrated image.}
	\label{fig:high-level-pipeline}
\end{figure*}
\end{abstract}
\section{Introduction}
\vspace*{-0.2cm}
3D reconstruction from image data is critically dependent on an accurate camera model, regardless of whether the approach is active or passive. Camera models based on intrinsic parameters, namely focal length, principal point, and distortion, enable inference of real-world coordinates from the image plane. Traditional calibration employs planar targets such as checkerboards or circle patterns imaged from multiple viewpoints, with parameter estimation performed by least-squares adjustment. Calibration quality depends on measurement count, viewpoint diversity, and image quality, and predominantly planar strategies can introduce high parameter correlations while insufficiently capturing variation along different rays \cite{luhmann-accuracy}.\\
This paper advances camera calibration for 3D reconstruction by proposing a flexible framework that uses scene-ray predictions and a robust icosahedral calibration target extending geometric information beyond traditional planar patterns. A generalized distortion model is evaluated using two novel ray-based metrics, \textit{intersection error} and \textit{reconstruction error}, alongside the conventional reprojection error, enabling a systematic comparison of camera models on metrics directly tailored to 3D reconstruction. On synthetic data with known ground-truth geometry, calibration stability is assessed via bootstrap resampling, providing confidence intervals for all error metrics~\cite{bootstrapping}. The results demonstrate the potential of three-dimensional calibration targets to achieve higher accuracy and calibration stability while offering greater flexibility in experimental design and practical deployment. In particular, our main contributions are:
\begin{itemize}
	\item \textbf{Optimized ring-based feature detection} for planar and 3D targets, combining border following, ellipse fitting, and sub-pixel center refinement using cross-ratio constraints and planar homologies.
	\item \textbf{Ray-based calibration metrics and experimental evaluation}, introducing intersection and reconstruction errors in object space alongside the known reprojection error, and using them to compare different camera models on real image data.
	\item \textbf{Novel 3D icosahedron calibration target} with ring grids on each face and combinatorial dot codes, plus algorithms to automatically detect faces, reconstruct the surface grids, and assign consistent 3D coordinates to all features.
	\item \textbf{Bootstrap-based evaluation on synthetic pinhole datasets}, using established and novel metrics to quantify calibration performance	variance, stability, and robustness across different 3D calibration targets.
\end{itemize}

\vspace*{-0.3cm}
\section{Related Work}
\vspace*{-0.2cm}
Camera calibration has a long history in photogrammetry~\cite{photogrammetry-history,luhmann-book}. The classical geometric model is the pinhole camera, and real-world compound lens systems are typically modeled as a parametric deviation from this ideal projection~\cite{brown-conrady,division-model,familiy-distortion-models,analytical-undistortion,segmentation-model,distortion-polynomial-inequalities,camera-pose-implicit-distortion}. When the camera deviates significantly from the pinhole model, as in wide-angle or fisheye configurations, more elaborate distortion models are required~\cite{kannala-brandt,review-wide-angle-camera,%
survey-wide-angle-image-rectification,overview-fish-eye-distortion,%
fish-eye-entrance-pupil}. More flexible formulations are also needed for unconventional camera types such as Scheimpflug cameras~\cite{review-scheimpflug}, plenoptic cameras~\cite{focused-plenoptic-camera,lifcal}, and catadioptric systems~\cite{towards-generic-camera,generic-cameras-single-center,%
smooth-models,non-parametric-structure-based,perfect-pinhole,%
review-wide-angle-camera}. In the most general formulations, the direction from which light is captured is directly encoded for each image location, giving rise to generalized camera models~\cite{generic-concept-calibration,towards-generic-camera,unifying-model,general-imaging-model,raxel-imaging-model,generic-cameras-single-center,parametric-general-model,smooth-models,10000-parameters}. Any calibration procedure requires correspondences between object points and their projections in the image. While such correspondences can be extracted from natural images~\cite{bundle-adjustment,targetless-camera-calibration,target-free-network,review-wide-angle-camera}, dedicated calibration targets provide higher accuracy because their feature points lie at known positions and can be robustly detected. High-precision features are most easily defined on planar targets, including patterns based on line intersections such as checkerboards~\cite{zhang-calibration}, deltille grids~\cite{deltille-grids}, and star patterns~\cite{10000-parameters,feature-extraction-reimagined}, as well as circle patterns whose detectable feature points are the circle centers~\cite{plane-based-calibration,circle-projection-model,need-accurate-pattern,concentric-circles-detection,collinear-points-pole-polar}. Feature identification within a target can be achieved through AprilTags~\cite{apriltag,feature-extraction-reimagined}, local feature constellations~\cite{robust-to-incomplete-pattern}, or patterns rendered on LCD screens~\cite{using-flat-monitors,speckle-pattern-on-display,feature-extraction-reimagined,generic-cameras-single-center}. Although accurate planar calibration targets are comparatively easy to manufacture, they tend to induce stronger correlations among the estimated parameters~\cite{luhmann-accuracy}, which motivates the use of true 3D calibration objects. These are commonly constructed by placing planar patterns onto the faces of a polyhedral structure augmented with markers for unambiguous face identification~\cite{rapid-calibration-by-one-image,deltille-grids}. Despite extensive prior work, a gap remains between generalized camera models and practical applications that support both planar and 3D targets with quantitative uncertainty estimates. This work addresses that gap by introducing a unified calibration framework combining optimized ring-based feature detection with a generalized camera model suitable for strongly distorted imaging systems. The framework is evaluated on synthetic and real data, covering metric comparison analysis, calibration target comparison, and derivation of statistically grounded uncertainty measures.

\section{Methodology}
\begin{figure}[t!]
	\centering
	\includegraphics[width=1\textwidth]{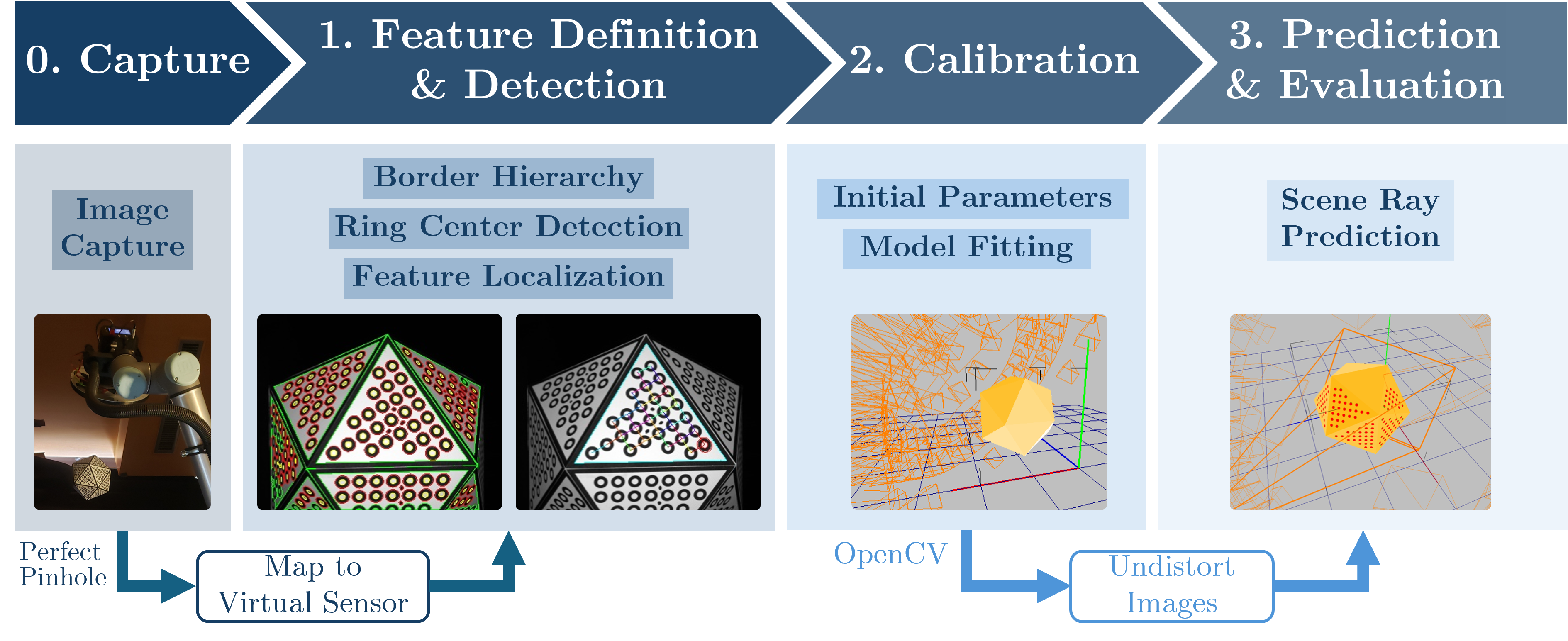}
	\caption{A flowchart representation of the proposed calibration pipeline.}
	\label{fig:flowchart}
\end{figure}
\subsection{Feature Definition and Detection}
\begin{figure}[!b]
	\centering
	\includegraphics[width=\textwidth]{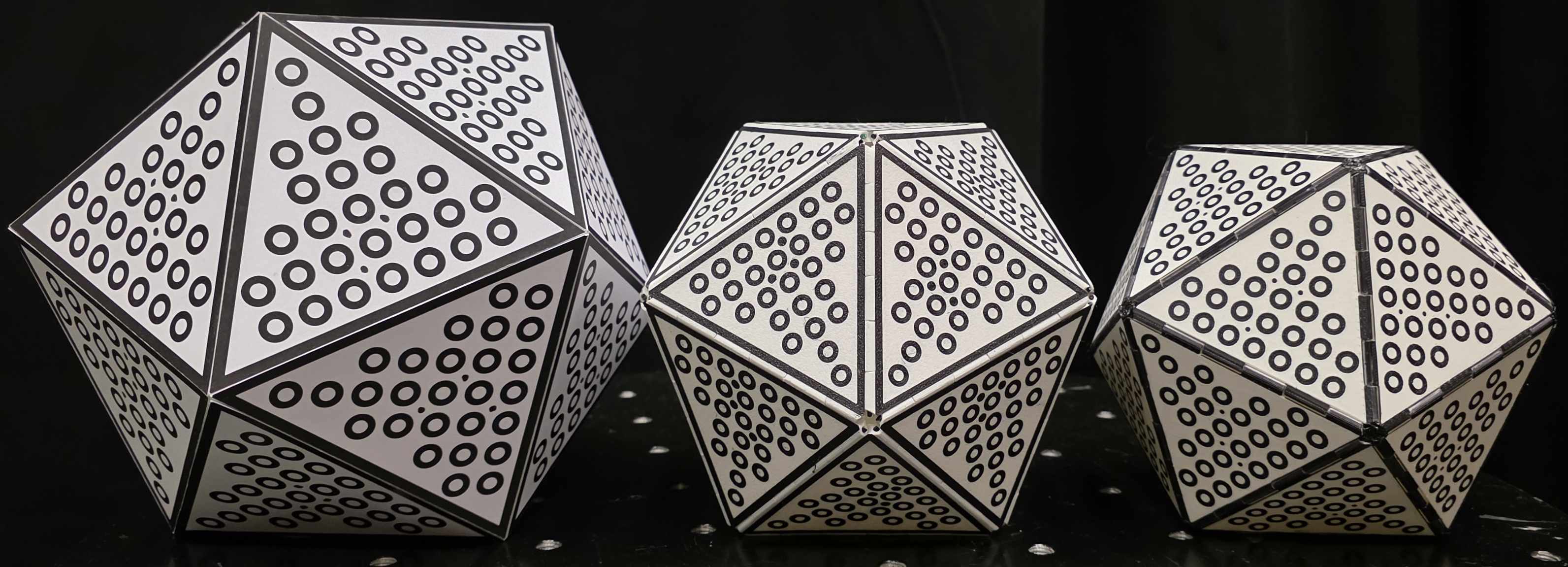}
	\caption[Constructed icosahedron targets]{3D-printed icosahedra. From left to right: MasterIco with adhered calibration patterns, ShinyIco with directly printed calibration patterns in shiny filament, and MattIco with directly printed calibration patterns in matte filament.}
	\label{fig:ico-target}
\end{figure}

\vspace*{-0.2cm}
The proposed calibration pipeline is summarized in the flowchart in Fig.~\ref{fig:flowchart}. Beginning with \textbf{Step~0 (Capture)}, images of the calibration target are acquired with the camera to be calibrated. The target may be either a 2D board or a 3D calibration object. In \textbf{Step~1 (Feature Definition \& Detection)}, feature points on the target are defined and detected, forming a correspondence dataset~$\mathcal{D}$ that relates scene points in 3D space to their projected image locations on the sensor. In \textbf{Step~2 (Calibration)}, a camera model is fitted to~$\mathcal{D}$ to estimate the calibration parameters. Finally, in \textbf{Step~3 (Prediction \& Evaluation)}, the calibrated model is used to predict scene rays, and prediction quality is evaluated using the reconstruction error, the intersection error, and the conventional reprojection error~\cite{bundle-adjustment}.\\
\begin{figure}[!b]
	\centering
	\begin{minipage}{.31\textwidth}
		\centering
		\includegraphics[width=\linewidth]{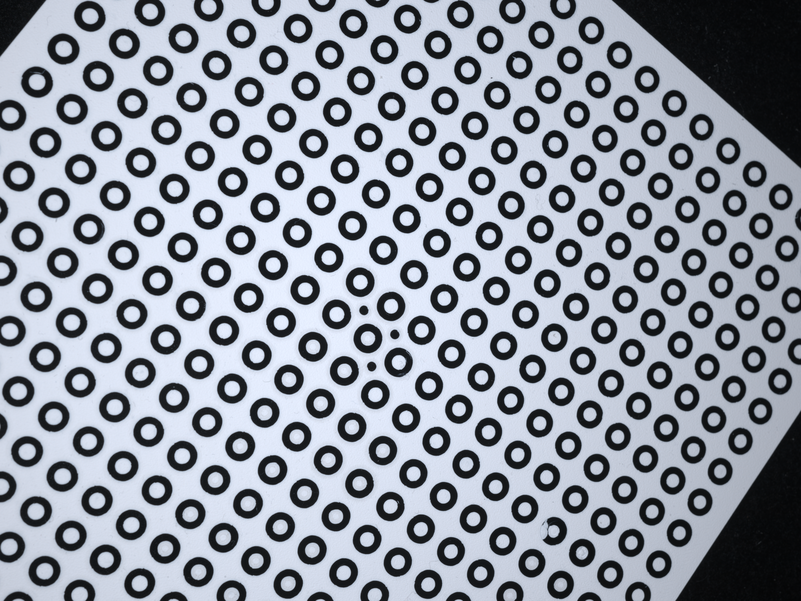}\\
		a)
	\end{minipage} 
	\begin{minipage}{.31\textwidth}
		\centering
		\includegraphics[width=\linewidth]{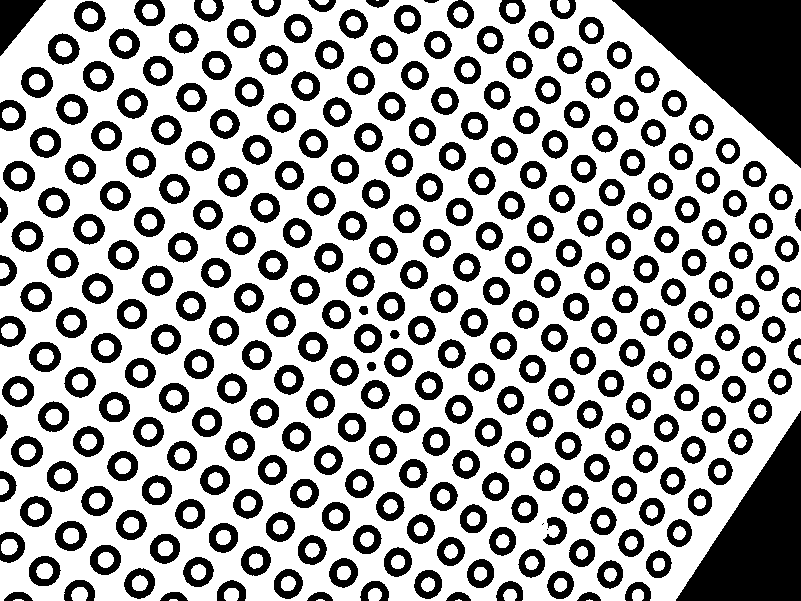}\\
		b)
	\end{minipage}
	\begin{minipage}{.31\textwidth}
		\centering
		\includegraphics[width=\linewidth]{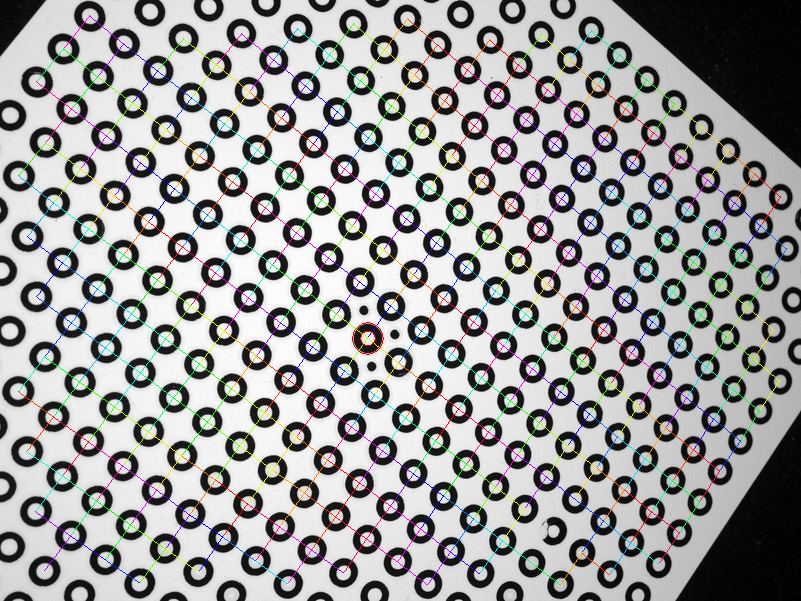}\\
		c)
	\end{minipage}
	\caption[Our ring board calibration target]{Example of a 2D calibration target: a) original image, b) binarized image, and c) automatically detected ringboard grid oriented around the marked central feature.}
	\label{fig:ringboard}
\end{figure}In \textbf{Step~1}, the correspondence dataset $\mathcal{D} = \{(\mathbf{u}_k,\mathbf{X}_{O,k}) : k = 1,\dots,n\}$ is constructed from $n$ calibration images. Features are defined as the centers of concentric circle pairs (Fig.~\ref{fig:ringboard}\,a), a well-established form of feature definition~\cite{concentric-circles-detection,collinear-points-pole-polar}. Binary masks are computed via Otsu thresholding~\cite{otsu-threshold}, and a border hierarchy is extracted via the Border Following algorithm~\cite{border-following}. Initial seed points are placed in white regions with a hierarchy depth of at least three, and are subsequently refined using the method of Jiang and Quan~\cite{concentric-circles-detection}.\\
The refinement is adapted to exploit the border hierarchy to compute line-to-ring intersections by projecting border elements onto the line. After filtering out improbable features, the surrounding borders of each detected feature are fit as ellipses and further constrained to satisfy a planar homology relationship~\cite{projective-geometry-book} using the Levenberg-Marquardt algorithm~\cite{levenberg,marquardt}. Grid localization identifies two neighboring features to define the grid axes and extends the full grid piecewise under an affine transformation which maps a local ring onto the unit circle. Three dot markers placed near the center of the target encode the grid origin and orientation unambiguously (Fig.~\ref{fig:ringboard}\,c). The final dataset~$\mathcal{D}$ assigns the grid coordinates as~$\mathbf{X}_O$ and the refined image positions as~$\mathbf{u}$.\\
The detection procedure extends naturally to a 3D calibration target. An icosahedron serves as the base geometry, with a grid of ring features placed on each face surrounded by a black border that separates features of adjacent faces using the border hierarchy (Fig.~\ref{fig:ico-target}). Planar grid coordinates are already determined from the detected grid shape. For face identification, six candidate grid positions are defined, of which exactly three are occupied by dot marks, yielding $\binom{6}{3} = 20$ distinct combinations that encode the 20 faces. The combination of grid coordinates and face identity uniquely determines the 3D position of each detected feature.
\subsection{Calibration}
\vspace*{-0.2cm}
With feature detection completed, \textbf{Step~2 (Calibration)} estimates the camera parameters by fitting the correspondence dataset to a camera model. The traditional model is based on the Brown distortion formulation~\cite{brown-conrady} as popularized by Zhang~\cite{zhang-calibration}, extending the linear pinhole projection with a parametric distortion model that captures the non-linear behavior of the lens. The following summary adopts the notation of Luhmann~\cite{luhmann-book}. In the linear pinhole model, the image projection $\mathbf{x}_i = (x_i, y_i)^\top$ of a scene point $\mathbf{X}_O = (X_O, Y_O, Z_O)^\top$ is obtained from the collinearity equations
\begin{align}
	x_i &= p_x - f_x
	\frac{r_{11}(X_O-t_X)+r_{21}(Y_O-t_Y)+r_{31}(Z_O-t_Z)}
	{r_{13}(X_O-t_X)+r_{23}(Y_O-t_Y)+r_{33}(Z_O-t_Z)},
	\nonumber \\
	y_i &= p_y - f_y
	\frac{r_{12}(X_O-t_X)+r_{22}(Y_O-t_Y)+r_{32}(Z_O-t_Z)}
	{r_{13}(X_O-t_X)+r_{23}(Y_O-t_Y)+r_{33}(Z_O-t_Z)},
	\label{eq:co_lin}
\end{align}
with extrinsic parameters, rotation $r_{\cdot\cdot}$ and translation $t_\cdot$, and intrinsic parameters, focal-length $f_\cdot$ and principal-point $p_\cdot$, grouped in vectors $\mathbf{f}$ and $\mathbf{p}$. Note that pixel coordinates $\mathbf{u} = \mathrm{diag}(1,-1)\cdot\mathbf{x}_i$ because the camera is oriented toward the negative $Z$-axis of the local camera coordinate system.\\

\begin{figure*}[!t]
	\centering
	\begin{minipage}{.45\textwidth}
		\centering
		\includegraphics[width=\linewidth]{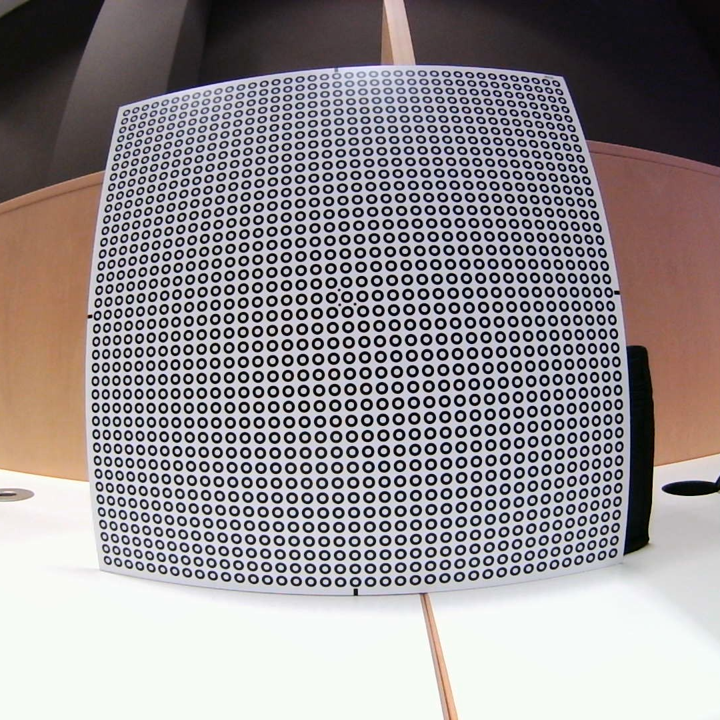}\\
		(a)
	\end{minipage} \hfill
	\begin{minipage}{.45\textwidth}
		\centering
		\includegraphics[width=\linewidth]{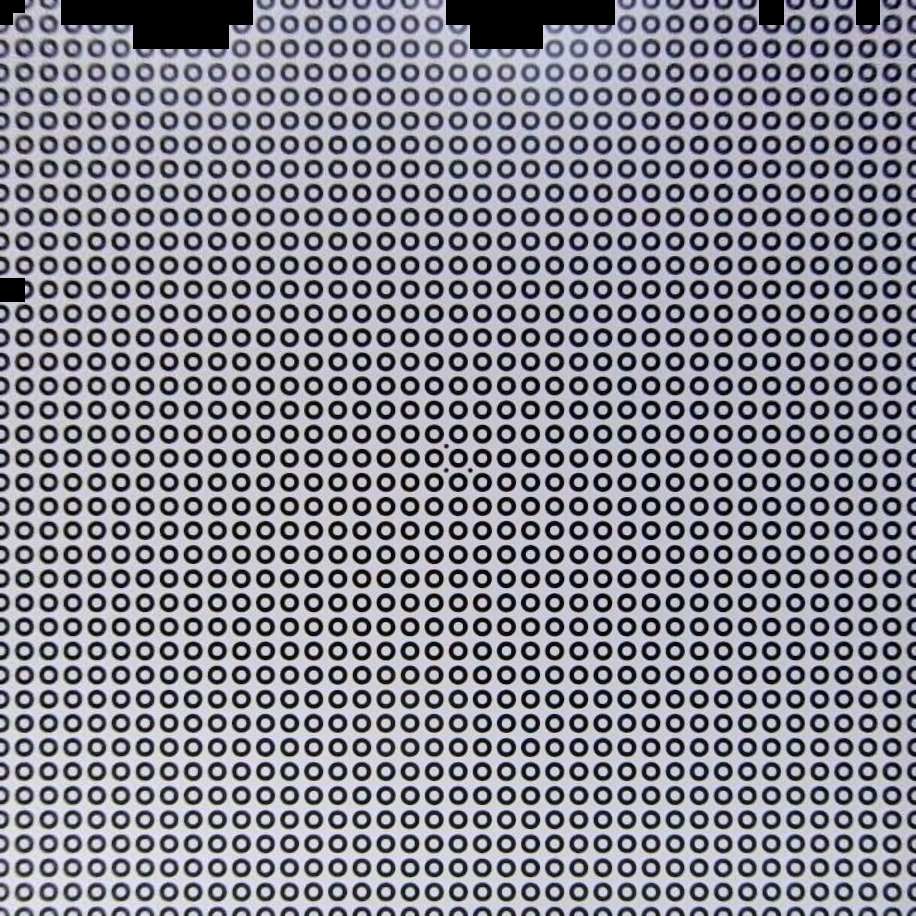}\\
		(b)
	\end{minipage}
	\caption[Virtual sensor mapping]{(a) Input image of the planar grid used to define the distortion-free virtual sensor. (b) Input image mapped onto the virtual sensor. The black regions block out areas with insufficient information for a reliable mapping.}
	\label{fig:virtual-sensor}
\end{figure*}

Lens distortion is modeled with a \textit{backward} formulation in which the distortion offset $\Delta$ is subtracted from the measured image coordinates~$\mathbf{x}_{ip}$~\cite{unified-distortion}. Working in normalized pinhole coordinates $\mathbf{x}_{pd} = \mathrm{diag}(\mathbf{f})^{-1}(\mathbf{x}_{ip}-\mathbf{p})$, the predicted undistorted position is $\bar{\mathbf{x}}_p = \mathbf{x}_{pd} - \Delta$, where the offset decomposes as $\Delta = \Delta^{rad} + \Delta^{tan}$. The radial term is
\begin{align*}
	\Delta^{rad} = K\,\mathbf{x}_{pd}, \quad
	K = k_1 r_{pd}^2 + k_2 r_{pd}^4 + k_3 r_{pd}^6, \quad
	r_{pd}^2 = \lVert\mathbf{x}_{pd}\rVert^2,
\end{align*}
and the tangential term, with parameters $p_1, p_2$, is defined as per Brown~\cite{brown-conrady}:
\begin{align*}
	\Delta_x^{tan} &= p_1(r_{pd}^2 + 2x_{pd}^2) + 2p_2\,x_{pd}\,y_{pd},
	\\
	\Delta_y^{tan} &= p_2(r_{pd}^2 + 2y_{pd}^2) + 2p_1\,x_{pd}\,y_{pd}.
\end{align*}
Initial parameters are obtained from homographies for planar targets~\cite{zhang-calibration} or via a direct linear transform for 3D features~\cite{dlt-derivation}, and are subsequently refined by the Levenberg-Marquardt algorithm~\cite{levenberg,marquardt}.\\
The flowchart in Fig.~\ref{fig:flowchart} highlights where the three evaluated pipelines diverge. In the traditional OpenCV model~\cite{opencv}, a \textit{forward} distortion formulation is used; a calibrated OpenCV camera must first generate an undistorted image from which scene rays can be predicted. For the virtual-camera model, the method of De~Boi et al.~\cite{perfect-pinhole} constructs a mapping to distortion-free images using Gaussian processes. Once images have been undistorted by this mapping, as illustrated in Figs.~\ref{fig:high-level-pipeline} and~\ref{fig:virtual-sensor}, the virtual camera is calibrated with the standard linear pinhole model.
\subsection{Prediction and Evaluation}
\vspace*{-0.2cm}
In \textbf{Step 3 (Prediction \& Evaluation)}, the calibrated camera model and its parameters are used for subsequent 3D reconstruction tasks. Before such a calibration is deployed, its performance must be evaluated. In the literature, the metric most commonly used to evaluate individual calibrations is the \textit{reprojection error}~\cite{bundle-adjustment}. However, the reprojection error alone is not fully adequate for assessing the accuracy of the proposed pipeline. Therefore, two additional metrics are introduced that are more directly related to 3D reconstruction: the \textit{intersection error} and the \textit{reconstruction error}.
\vspace*{-0.2cm}
\paragraph{\textbf{\textit{Reprojection Error}.}} The \textit{reprojection error} is defined as $\mathbf e_{\text{rep}} = \mathbf u_{\text{pred}} - \mathbf u_{\text{det}}$, where $\mathbf u_{\text{pred}}$ is the image location of a scene feature predicted by the calibrated camera model and $\mathbf u_{\text{det}}$ is the measured location thereof. The individual reprojection errors are aggregated as a root-mean squared error
\begin{align*}
	\mathrm{RMSE}_{\mathrm{rep}} &= \sqrt{\frac{1}{n} \sum_{k=1}^n \left\lVert \mathbf e_{\mathrm{rep}}^k \right\rVert_2^2 } \, ,
\end{align*}
where $n$ is the total number of detected features~\cite{rmse}. However, this metric is closely related to the optimization objective, so a lower value does not necessarily imply a better calibration. Moreover, in most applications the primary interest lies in 3D accuracy in object space rather than image-plane accuracy, making the reprojection error only an indirect indicator of calibration quality for reconstruction tasks.
\paragraph{\textbf{Intersection Error.}} To overcome the limitations of the reprojection error and to place greater emphasis on the accuracy of 3D reconstruction, an additional metric, termed the \textit{intersection error}, is introduced. Given an image point, the camera model is used to predict the corresponding scene ray, which represents the path of light from a point in the scene toward the camera sensor. This ray-tracing operation is supported by most camera models, since they are based on the assumption that light travels along straight lines. The predicted scene ray is then intersected with the known geometry of the calibration target to obtain the predicted 3D location of the feature point $\mathbf{X}^{\text{pred}}$, defining the intersection error as $\mathbf{e}_{\mathrm{int}} = \mathbf{X}^{\text{pred}} - \mathbf{X}_O \, $. This error measure evaluates accuracy directly in object space and is expressed in metric units.
\paragraph{\textbf{Reconstruction Error.}} In contrast to the intersection error, which exploits the known target geometry, this metric estimates the feature point positions using only the predicted scene rays. For each feature point on the target, all corresponding rays from the different camera poses in the calibration dataset are collected. The 3D feature position is then computed as the point that is, in a least-squares sense, closest to all of these rays. Each ray is represented by a direction vector $\mathbf d_k$ and a point $\mathbf s_k$ on the ray. For an arbitrary 3D point $\mathbf X_r$, the perpendicular distance to ray $k$ equals $\lVert \mathbf v_k \rVert_2$~\cite{wolfram-alpha-point-line-distance}, where
\begin{align} \label{eq:reco-point}
	\mathbf v_k &= \mathbf d_k \times (\mathbf s_k - \mathbf X_r) \, .
\end{align}
The reconstructed 3D feature position is obtained by minimizing the sum of squared distances to all rays based on Equation~\eqref{eq:reco-point}. In contrast to the intersection error, the reconstruction error yields a single error value per feature point on the target and can average out individual ray errors by incorporating multiple views. As such, it represents a lower bound on the achievable reconstruction accuracy when a given calibrated camera model is employed in a 3D reconstruction task.\\

\vspace*{-0.2cm}
Using these newly introduced metrics enables evaluation of 3D reconstruction tasks in a way that aligns more closely with the underlying problem, without relying on the reprojection error, an indirect measure. The following experiments in Section \ref{sec:results} report all three metrics for comparison. While the reprojection, intersection, and reconstruction errors quantify different aspects of geometry reconstruction accuracy in object space, the use of synthetic data provides the additional opportunity to evaluate the estimated camera poses against known ground-truth poses.
\vspace*{-0.2cm}
\paragraph{\textbf{Pose Error.}} In the case of synthetically generated data, the exact capture poses for each image are known. Each image is generated by placing a virtual pinhole camera at a specified pose in 3D space and tracing a ray through every pixel of the image. Whenever a ray intersects the calibration target, the pixel intensity is set according to the target texture (e.g., black or white). The resulting capture pose $\mathbf P_{gt}$ serves as ground truth and can be compared to the calibrated pose $\mathbf P$. Due to the simulation process, this comparison is restricted to calibrated camera models that are also based on the pinhole model. For each image used during calibration, the pose error is computed as the distance between the estimated pose and its corresponding ground-truth pose, $\mathbf e_{\mathrm{pose}} = \lVert \mathbf P - \mathbf P_{gt} \rVert \, $. \vspace*{-0.3cm}
\section{Results} \label{sec:results}
\begin{figure}[!t]
	\begin{minipage}{.47\textwidth}
			\centering
			\includegraphics[width=\linewidth]{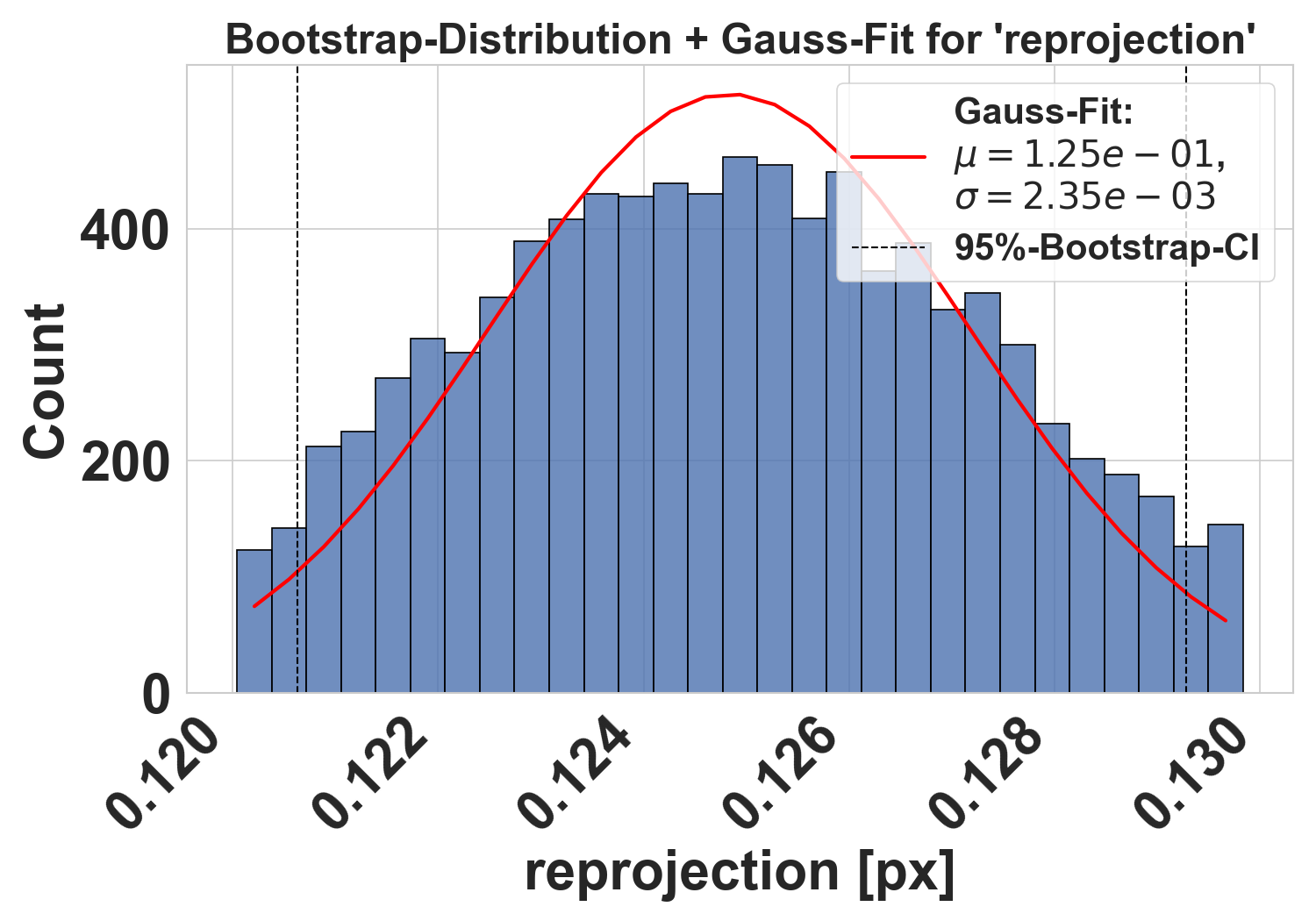} \\ \hspace*{0.6cm} Planar Ringboard 
		\end{minipage} \hfill
	\begin{minipage}{.47\textwidth}
			\centering
			\includegraphics[width=\linewidth]{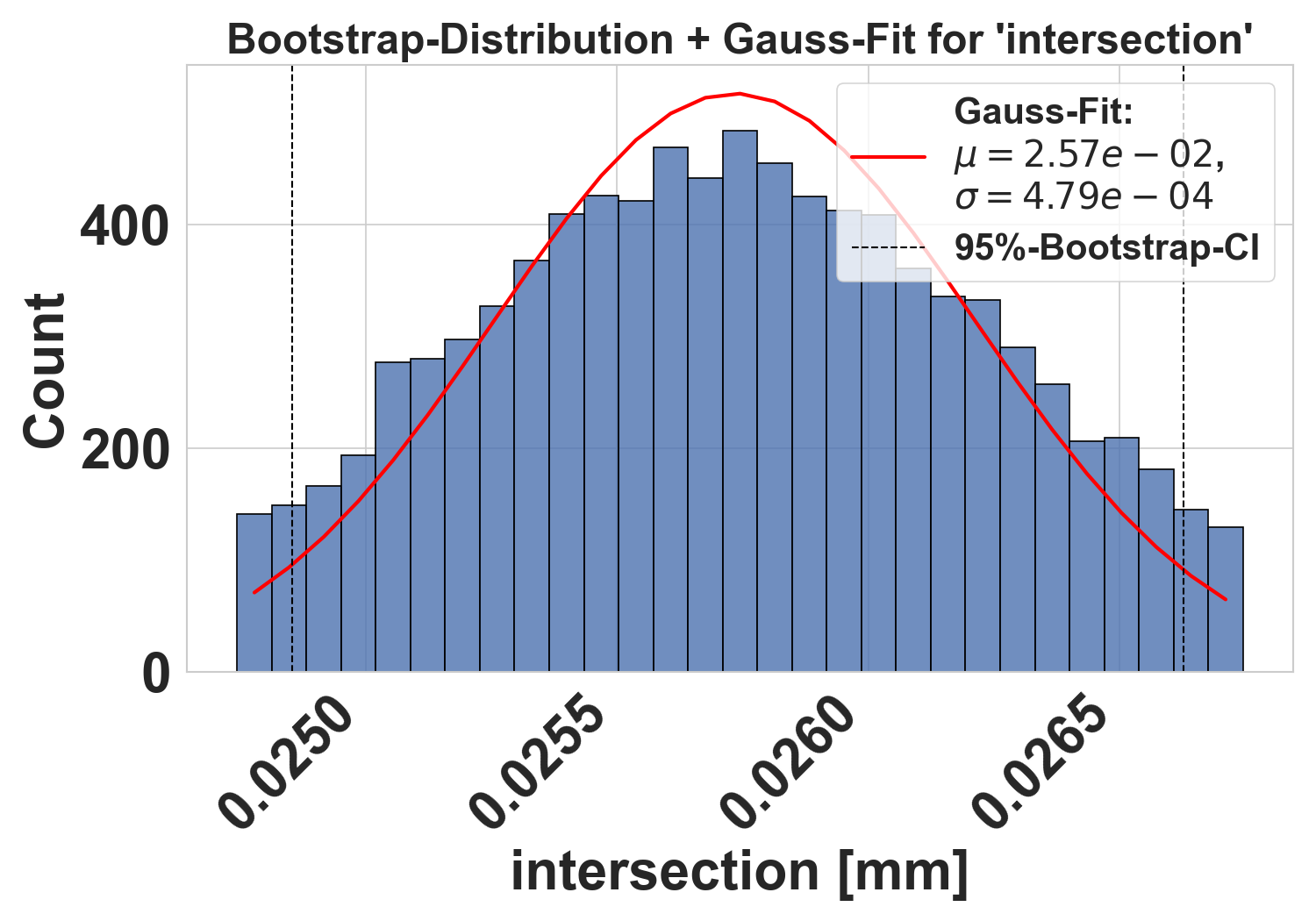} \\ 	\hspace*{0.6cm} Icosahedron
		\end{minipage}
	\caption{Histograms, fitted Gaussian curves, and confidence intervals of the reprojection error for the planar ringboard and the intersection error for the icosahedron.}
	\label{fig:histograms}
	\vspace*{-0.4cm}
\end{figure}

\vspace*{-0.2cm}
The experimental evaluation comprises two complementary parts. First, synthetic image data with known ground-truth camera poses and scene geometry are generated in order to systematically analyze how different calibration targets behave under idealized conditions. In this synthetic setting, the availability of ground-truth poses enables an evaluation of pose error. Due to the generation process, the synthetic evaluation is restricted to the Luhmann model and is supported by a bootstrap procedure that quantifies the expected variance and robustness of the resulting calibrations. Second, a Phase One iXG camera with a 100 megapixel sensor is calibrated with the proposed pipelines and 5 calibration targets. These real-image datasets are used to compare camera models on practical data using the reprojection, intersection, and reconstruction errors.
\begin{table}[!b]
	\scriptsize
	\caption{Length (L), width (W), height (H), concentric circle diameter (D) and center distances (C) of the calibration targets in millimeters [mm].}
	\label{table:calibration_objects}	
	\setlength\tabcolsep{0pt}
	\begin{tabular*}{\columnwidth}{@{\extracolsep{\fill}}cccccc@{}}
		\toprule
		\multirow{2}{*}{\shortstack{\textit{Calibration}\\\textit{Object}}}& \multirow{2}{*}{\textit{MasterIco}} & \multirow{2}{*}{\textit{MattIco}} & \multirow{2}{*}{\textit{ShinyIco}} & \multirow{2}{*}{\textit{AluBoard}}  & \multirow{2}{*}{\textit{3DBoard}} \\ 
		&&&&&\\
		\midrule
		\multirow{2}{*}{L$\times$W$\times$H }& \multirow{2}{*}{20$\times$22.5$\times$25} & \multirow{2}{*}{15.1$\times$16.9$\times$18.5}&\multirow{2}{*}{16.1$\times$17.9$\times$20} &\multirow{2}{*}{40$\times$40$\times$3}& \multirow{2}{*}{22$\times$22$\times$5}\\
		&&&&&\\
		\multirow{2}{*}{D}& \multirow{2}{*}{10.5} & \multirow{2}{*}{7}&\multirow{2}{*}{7} &\multirow{2}{*}{8}& \multirow{2}{*}{7}\\
		&&&&&\\
		\multirow{2}{*}{C}& \multirow{2}{*}{13.2} & \multirow{2}{*}{10}&\multirow{2}{*}{10} &\multirow{2}{*}{10}& \multirow{2}{*}{10}\\
		&&&&&\\
		\bottomrule
	\end{tabular*}
	\vspace*{-0.6cm}
\end{table}
\paragraph{\textbf{Synthetic image generation.}} To effectively compare the accuracy of the proposed 3D icosahedron calibration target with conventional planar ringboard calibration targets, synthetic datasets were generated using a ray-tracing-based rendering approach. In this synthetic setting the exact intrinsic and extrinsic parameters are known, which enables a direct comparison between estimated and ground-truth values. Approximately 200 synthetic images were rendered from distinct camera poses at a focus distance of 40~centimeters between the virtual camera and the calibration target. The camera poses were distributed nearly uniformly on a hemisphere around the calibration target, with all cameras oriented toward the origin \cite{swinbank2006fibonacci}.
\paragraph{\textbf{Bootstrap Procedure.}}
Calibration stability on the synthetic datasets is assessed via bootstrap resampling~\cite{bootstrapping}. From the full pool of approximately 200 rendered images, $B = 10{,}000$ bootstrap samples are drawn with replacement, each of size $N \in \{20, 50, 120\}$ images. For each sample, the complete Luhmann pipeline is executed and all four error metrics are computed. The resulting distributions are summarized by a fitted Gaussian, yielding mean~$\mu$ and standard deviation~$\sigma$. The 95\,\% confidence intervals reported in Table~\ref{table:results} are derived from these distributions and visualized in Fig. \ref{fig:histograms}.
\paragraph{\textbf{Calibration targets.}}
The calibration target setup is based on 3D-printed designs and comprises five individual objects in total. The first object is manufactured as a single-color print, with the feature patterns applied by hand (\textit{MasterIco}). This prototype is sufficient to test the overall concept, but the manual application of the patterns can introduce geometric inaccuracies. To obtain more precise feature layouts, a second icosahedral target (\textit{ShinyIco}) is produced using a dual-filament 3D printer, in which the calibration pattern is printed directly onto the surface. The shiny finish of this print, however, generates specular highlights in the captured images. A third variant (\textit{MattIco}) is therefore manufactured using a matte filament, which provides similarly accurate printed calibration patterns while eliminating disturbing highlights in the captured images. The three icosahedral targets were printed using the Bambu Lab P1S multi-material printer and are shown in Fig.~\ref{fig:ico-target}.
In addition to these 3D targets, two planar calibration targets are used. The first consists of an aluminum composite sheet onto which the ringboard calibration pattern is conventionally printed (\textit{AluBoard}), and the second is a dual-filament 3D-printed board (\textit{3DBoard}), as visualized in Figs.~\ref{fig:ringboard} and~\ref{fig:virtual-sensor}. Together, these targets cover both traditional planar configurations and 3D calibration targets. Table~\ref{table:calibration_objects} summarizes the dimensions of all calibration targets. For each object, length, width, and height are measured by placing the object on a flat face, recording the height perpendicular to that face, and using the in-plane extents as length and width. In addition, Table~\ref{table:calibration_objects} reports the diameters of the concentric circles and the center-to-center distances between adjacent features of the calibration patterns.
\paragraph{\textbf{Real-image acquisition.}}
Real data were acquired with the autonomous 3D scanning system CultArm3D~\cite{SantosPedroFraunhoferIGD2020A3MD}. The system combines a six-axis robotic arm, a motorized turntable, a ring light, and a PhaseOne iXG 100 megapixel camera with a 72\,mm lens. During acquisition a calibration target is placed on the turntable and digitized fully autonomously. The robot and turntable execute poses adapted to the target size so that all surface regions are imaged with sufficient overlap. This procedure yields about 30 images for planar targets and about 90 images for icosahedral targets. For each calibration object in Table~\ref{table:calibration_objects}, 12 independent image sets are captured, resulting in 60 image datasets.
\begin{table*}[!b]
	\scriptsize
	\caption{Gaussian fit parameters (mean $\mu$ and standard deviation $\sigma$) of reconstruction (Recon.), intersection (Inter.), reprojection (Repro.), and pose errors (Pose) for simulated planar board and icosahedron calibration targets at a focus distance of 40~cm. Results are shown for different numbers of calibration images (20, 50, 120) including the confidence intervals (-CI). Values marked with an asterisk ($^{*}$) are discussed in the text.}
	\label{table:results}
	\setlength\tabcolsep{0pt}
	\begin{tabular*}{\textwidth}{@{\extracolsep{\fill}}lcccccccc|cccccccc@{}}
		\toprule
		\multirow{2}{*}{\textit{Object}} & \multicolumn{8}{c|}{\multirow{2}{*}{Planar Ringboard}} &  \multicolumn{8}{c}{\multirow{2}{*}{Icosahedron}} \\ 
		&&&&&&&&&&&&&&&&\\
		\midrule
		& \multicolumn{2}{c}{\multirow{3}{*}{\shortstack{Recon.\\ $[10^{-2}\,\mathrm{mm}]$}}}
		& \multicolumn{2}{c}{\multirow{3}{*}{\shortstack{Inter.\\ $[10^{-2}\,\mathrm{mm}]$}}}
		& \multicolumn{2}{c}{\multirow{3}{*}{\shortstack{Repro.\\ $[10^{-1}\,\mathrm{px}]$}}}
		& \multicolumn{2}{c|}{\multirow{3}{*}{\shortstack{Pose\\ $[10^{-1}\,\mathrm{mm}]$}}}
		& \multicolumn{2}{c}{\multirow{3}{*}{\shortstack{Recon.\\ $[10^{-2}\,\mathrm{mm}]$}}}
		& \multicolumn{2}{c}{\multirow{3}{*}{\shortstack{Inter.\\ $[10^{-2}\,\mathrm{mm}]$}}}
		& \multicolumn{2}{c}{\multirow{3}{*}{\shortstack{Repro.\\ $[10^{-1}\,\mathrm{px}]$}}}
		& \multicolumn{2}{c}{\multirow{3}{*}{\shortstack{Pose\\ $[10^{-1}\,\mathrm{mm}]$}}}\\
		&&&&&&&&&&&&&&&&\\
		& \multirow{2}{*}{$\mu$} & \multirow{2}{*}{$\sigma$} & \multirow{2}{*}{$\mu$} & \multirow{2}{*}{$\sigma$} & \multirow{2}{*}{$\mu$} & \multirow{2}{*}{$\sigma$} & \multirow{2}{*}{$\mu$} & \multirow{2}{*}{$\sigma$} & \multirow{2}{*}{$\mu$} & \multirow{2}{*}{$\sigma$} & \multirow{2}{*}{$\mu$} & \multirow{2}{*}{$\sigma$} & \multirow{2}{*}{$\mu$} & \multirow{2}{*}{$\sigma$} & \multirow{2}{*}{$\mu$} & \multirow{2}{*}{$\sigma$} \\
		&&&&&&&&&&&&&&&&\\
		\midrule
		\multirow{2}{*}{20} &
		\multirow{2}{*}{0.83} & \multirow{2}{*}{0.05} &
		\multirow{2}{*}{4.27} & \multirow{2}{*}{0.48} &
		\multirow{2}{*}{1.24} & \multirow{2}{*}{0.06} &
		\multirow{2}{*}{1.44*} & \multirow{2}{*}{0.87*} \textnormal{ }&
		\multirow{2}{*}{1.67} & \multirow{2}{*}{0.15} &
		\multirow{2}{*}{2.56} & \multirow{2}{*}{0.12} &
		\multirow{2}{*}{1.34} & \multirow{2}{*}{0.04} &
		\multirow{2}{*}{0.93*} & \multirow{2}{*}{0.30*} \\
		&&&&&&&&&&&&&&&&\\
		\multirow{2}{*}{CI-20} &
		\multicolumn{2}{l}{\multirow{2}{*}{$[0.74 \text{-} 0.94]$}}&
		\multicolumn{2}{l}{\multirow{2}{*}{$[0.34 \text{-} 0.52]$}}&
		\multicolumn{2}{l}{\multirow{2}{*}{$[1.14 \text{-} 1.35]$}}&
		\multicolumn{2}{l|}{\multirow{2}{*}{$[0.31 \text{-} 3.45]$}}&
		\multicolumn{2}{l}{\multirow{2}{*}{$[1.47 \text{-} 2.04]$}}&
		\multicolumn{2}{l}{\multirow{2}{*}{$[2.35 \text{-} 2.79]$}}&
		\multicolumn{2}{l}{\multirow{2}{*}{$[1.28 \text{-} 1.41]$}}&
		\multicolumn{2}{l}{\multirow{2}{*}{$[0.56 \text{-} 1.63]$}} \\
		&&&&&&&&&&&&&&&&\\ \midrule
		\multirow{2}{*}{50} &
		\multirow{2}{*}{0.55} & \multirow{2}{*}{0.03} &
		\multirow{2}{*}{4.30} & \multirow{2}{*}{0.31} &
		\multirow{2}{*}{1.25} & \multirow{2}{*}{0.04} &
		\multirow{2}{*}{0.87} & \multirow{2}{*}{0.51} \textnormal{ }&
		\multirow{2}{*}{1.30} & \multirow{2}{*}{0.07} &
		\multirow{2}{*}{2.57} & \multirow{2}{*}{0.07} &
		\multirow{2}{*}{1.35} & \multirow{2}{*}{0.02} &
		\multirow{2}{*}{0.82} & \multirow{2}{*}{0.21} \\
		&&&&&&&&&&&&&&&&\\
		\multirow{2}{*}{CI-50} &
		\multicolumn{2}{l}{\multirow{2}{*}{$[0.50 \text{-} 0.61]$}}&
		\multicolumn{2}{l}{\multirow{2}{*}{$[3.74 \text{-} 4.86]$}}&
		\multicolumn{2}{l}{\multirow{2}{*}{$[1.18 \text{-} 1.32]$}}&
		\multicolumn{2}{l|}{\multirow{2}{*}{$[0.21 \text{-} 2.00]$}}&
		\multicolumn{2}{l}{\multirow{2}{*}{$[1.19 \text{-} 1.45]$}}&
		\multicolumn{2}{l}{\multirow{2}{*}{$[2.44 \text{-} 2.71]$}}&
		\multicolumn{2}{l}{\multirow{2}{*}{$[1.30 \text{-} 1.39]$}}&
		\multicolumn{2}{l}{\multirow{2}{*}{$[0.56 \text{-} 1.30]$}} \\
		&&&&&&&&&&&&&&&&\\ \midrule
		\multirow{2}{*}{120} &
		\multirow{2}{*}{0.40*} & \multirow{2}{*}{0.02*} &
		\multirow{2}{*}{4.32*} & \multirow{2}{*}{0.20*} &
		\multirow{2}{*}{1.25*} & \multirow{2}{*}{0.02*} &
		\multirow{2}{*}{0.58} & \multirow{2}{*}{0.31*} \textnormal{ }&
		\multirow{2}{*}{1.10*} & \multirow{2}{*}{0.03*} &
		\multirow{2}{*}{2.57*} & \multirow{2}{*}{0.05*} &
		\multirow{2}{*}{1.35*} & \multirow{2}{*}{0.01*} &
		\multirow{2}{*}{0.78} & \multirow{2}{*}{0.15*} \\
		&&&&&&&&&&&&&&&&\\
		\multirow{2}{*}{CI-120} &
		\multicolumn{2}{l}{\multirow{2}{*}{$[0.37 \text{-} 0.43]$}}&
		\multicolumn{2}{l}{\multirow{2}{*}{$[3.96 \text{-} 4.68]$}}&
		\multicolumn{2}{l}{\multirow{2}{*}{$[1.21 \text{-} 1.29]$}}&
		\multicolumn{2}{l|}{\multirow{2}{*}{$[0.17 \text{-} 1.28]$}}&
		\multicolumn{2}{l}{\multirow{2}{*}{$[1.05 \text{-} 1.16]$}}&
		\multicolumn{2}{l}{\multirow{2}{*}{$[2.49 \text{-} 2.66]$}}&
		\multicolumn{2}{l}{\multirow{2}{*}{$[1.32 \text{-} 1.37]$}}&
		\multicolumn{2}{l}{\multirow{2}{*}{$[0.58 \text{-} 1.09]$}} \\
		&&&&&&&&&&&&&&&&\\
		\bottomrule
	\end{tabular*}
\end{table*}
\paragraph{\textbf{Simulation study.}}
Table~\ref{table:results} summarizes the synthetic evaluation. Across all metrics and both targets, increasing the number of calibration images consistently reduced the standard deviation, confirming that more diverse observations improve numerical stability.
Considering mean values alone, the \textbf{reprojection error} suggested a slight advantage for the planar board ($1.25\times10^{-1}$\,px vs.\ $1.35\times10^{-1}$\,px at 120 images), yet the icosahedron exhibited lower standard deviation at all image counts ($\sigma=0.01\times10^{-1}$\,px vs.\ $0.02\times10^{-1}$\,px), indicating more stable reprojection performance. The \textbf{intersection error} showed the opposite mean-value ordering: the icosahedron achieved roughly 40\,\% lower mean ($2.57\times10^{-2}$\,mm vs.\ $4.32\times10^{-2}$\,mm) and a fourfold lower standard deviation ($\sigma=0.05\times10^{-2}$\,mm vs.\ $0.20\times10^{-2}$\,mm at 120 images). These contrasting results illustrate that reprojection error alone can be misleading for reconstruction-oriented applications. The \textbf{reconstruction error} favored the planar board in both mean and standard deviation ($0.40\times10^{-2}$\,mm, $\sigma=0.02\times10^{-2}$\,mm vs.\ $1.10\times10^{-2}$\,mm, $\sigma=0.03\times10^{-2}$\,mm at 120 images), as each icosahedron feature was supported by fewer rays due to face occlusion across views, limiting multi-view averaging. The \textbf{pose error} further underscored the icosahedron's stability advantage: at 20 images, it achieved a mean of $0.93\times10^{-1}$\,mm with $\sigma=0.30\times10^{-1}$\,mm, while the planar board yielded a mean of $1.44\times10^{-1}$\,mm with a nearly threefold larger standard deviation of $\sigma=0.87\times10^{-1}$\,mm. At 120 images, the standard deviation of the icosahedron remained roughly twofold lower ($\sigma=0.15\times10^{-1}$\,mm vs.\ $0.31\times10^{-1}$\,mm).\\
Overall, mean values alone present an incomplete picture: the planar board achieved lower mean reprojection and reconstruction errors, yet the icosahedron delivered approximately 40\,\% lower mean intersection error alongside lower standard deviation for the intersection, reprojection, and pose errors. Calibration stability must therefore be considered alongside mean accuracy, and the choice of target should be guided by the primary metric of the intended application.
\paragraph{\textbf{Real-world evaluation.}}
\begin{figure*}[!t]
	\centering
    \includegraphics[width=\textwidth]{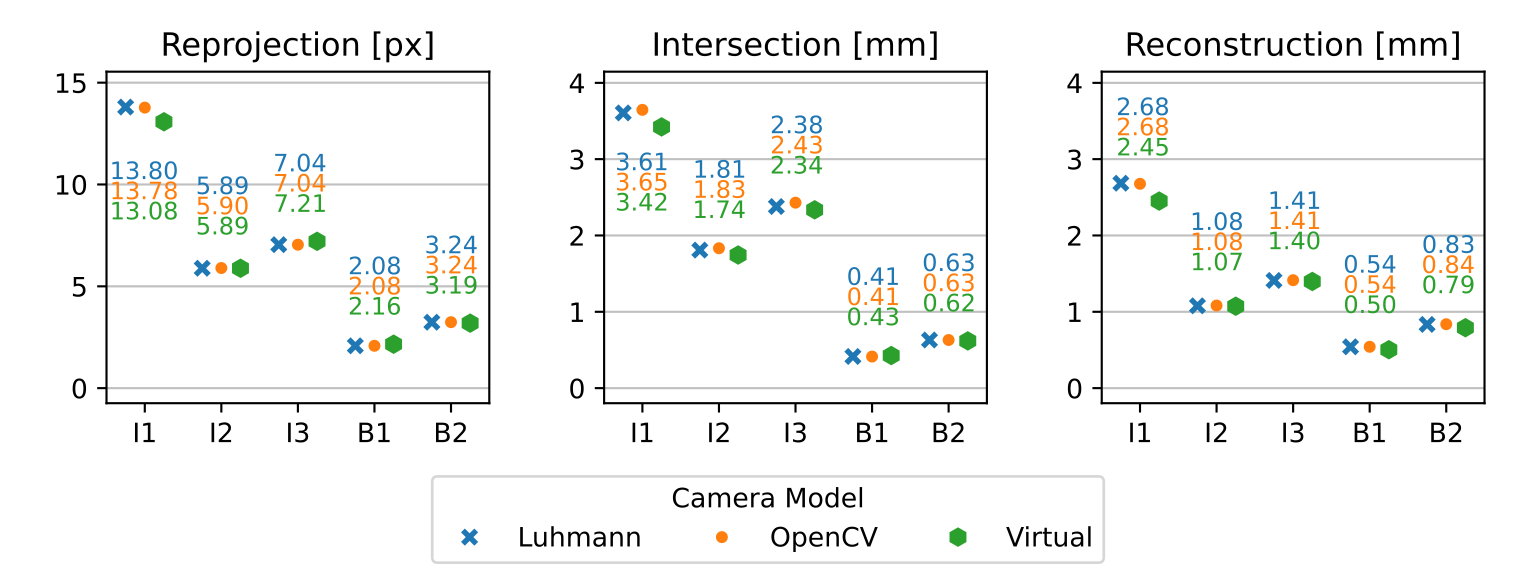}
	\caption[Evaluation of calibration targets]{Comparison of constructed calibration targets fitted with Luhmann, OpenCV, and virtual camera models. Shown are median values of 12 calibration runs for I1: MasterIco, I2: ShinyIco, I3: MattIco, B1: AluBoard, and B2: 3DBoard.}
	\label{fig:target_boxplot}
	\vspace*{-0.6cm}
\end{figure*}
\vspace*{-0.2cm}
Fig.~\ref{fig:target_boxplot} compares all calibration targets on the Phase One iXG camera. The planar targets (B1: AluBoard, B2: 3DBoard) achieved the lowest errors, with intersection errors around $0.4$\,mm (B1) and $0.6$\,mm (B2) and reconstruction errors of $0.5$\,mm (B1) and $0.8$\,mm (B2).
Among the icosahedral targets, MasterIco (I1) performed worst, which is attributed to the manually applied patterns; its intersection error exceeds $3$\,mm and its reconstruction error is approximately $2.7$\,mm. The best performing 3D target was ShinyIco (I2), achieving intersection errors between $1.74$\,mm and $1.83$\,mm and reconstruction errors of around $1.08$\,mm, an increase of two to four times compared to the best performing planar target.
The reprojection error showed a similar situation. However, the ratios between these errors were different. The reprojection error exhibited by ShinyIco (I2) was twice as large as that of the similarly manufactured 3DBoard (B2), though their reconstruction error only showed a difference of roughly 25\,\%, indicating that these targets performed more similarly than predicted by the reprojection error alone.

\section{Conclusion and Future Work}
\vspace*{-0.2cm}
This work introduced a calibration framework that combines ring-based feature detection on planar and 3D targets with ray-based evaluation metrics. Experiments on synthetic and real data demonstrated that intersection and reconstruction errors provide a more direct and informative assessment of calibration quality than the reprojection error, which can be misleading for 3D reconstruction tasks. \textbf{On synthetic data}, the icosahedral target achieved a roughly 40\,\% lower mean intersection error than planar boards and exhibited significantly more stable calibration results for intersection and pose estimation, with a fourfold and threefold reduction in standard deviation, respectively. This stability stemmed from the richer feature geometry provided by the icosahedron's 3D form, which reduced parameter correlations and yielded more consistent parameter estimates, particularly when fewer calibration images were available. Planar boards, by contrast, achieved better mean reconstruction and reprojection errors. The choice of calibration target should therefore be guided by the primary metric of the intended application. \textbf{In real-world experiments}, high-quality planar ringboards achieved the smallest errors across all metrics, while icosahedral targets exhibited higher errors attributable to geometric inaccuracies introduced during 3D printing. Closing this gap requires tighter fabrication tolerances, and quantifying manufacturing error is an important direction for future work. Still, the icosahedron target provided a practical flexibility that is difficult to achieve with planar targets, as it allowed captures from any angle. The virtual camera model yielded small but consistent improvements for planar targets in both synthetic and real-world evaluations.

\paragraph{\textbf{Future Work.}}
\vspace*{-0.2cm}
Future efforts will integrate ray-based metrics directly into bundle adjustment, improve the manufacturing precision of 3D targets to close the gap between synthetic and real-world performance, and investigate the use of other, more elaborate generalized models in a full 3D reconstruction pipeline.

%
%
%
\bibliographystyle{splncs04}
\bibliography{references}

\begin{thebibliography}{10}
\providecommand{\url}[1]{\texttt{#1}}
\providecommand{\urlprefix}{URL }
\providecommand{\doi}[1]{https://doi.org/#1}

\bibitem{dlt-derivation}
Abdel-Aziz, Y.I., Karara, H.M.: Direct linear transformation from comparator
  coordinates into object space coordinates in close-range photogrammetry.
  Photogrammetric Engineering \& Remote Sensing  \textbf{81}(2),  103--107 (2
  2015). \doi{10.14358/PERS.81.2.103}

\bibitem{targetless-camera-calibration}
Barazzetti, L., Mussio, L., Remondino, F., Scaioni, M.: Targetless camera
  calibration. International Archives of the Photogrammetry, Remote Sensing and
  Spatial Information Sciences  \textbf{XXXVIII-5/W16} (09 2012).
  \doi{10.5194/isprsarchives-XXXVIII-5-W16-335-2011}

\bibitem{opencv}
Bradski, G.: {The OpenCV Library}. Dr. Dobb's Journal of Software Tools  (2000)

\bibitem{brown-conrady}
Brown, D.C.: Decentering {Distortion} of {Lenses}. Photometric Engineering
  \textbf{32}(3),  444--462 (1966)

\bibitem{non-parametric-structure-based}
Camposeco, F., Sattler, T., Pollefeys, M.: Non-parametric structure-based
  calibration of radially symmetric cameras. In: International Conference on
  Computer Vision. pp. 2192--2200 (2015). \doi{10.1109/ICCV.2015.253}

\bibitem{photogrammetry-history}
Clarke, T.A., Fryer, J.G.: The development of camera calibration methods and
  models. The Photogrammetric Record  \textbf{16} (1998).
  \doi{10.1111/0031-868X.00113}

\bibitem{perfect-pinhole}
De~Boi, I., Pathak, S., Oliveira, M., Penne, R.: How to turn your camera into a
  perfect pinhole model. In: Progress in Pattern Recognition, Image Analysis,
  Computer Vision, and Applications. pp. 90--107 (2023)

\bibitem{generic-cameras-single-center}
Dunne, A.K., Mallon, J., Whelan, P.F.: Efficient generic calibration method for
  general cameras with single centre of projection. In: International
  Conference on Computer Vision. pp.~1--8 (2007).
  \doi{10.1109/ICCV.2007.4408990}

\bibitem{survey-wide-angle-image-rectification}
Fan, J., Zhang, J., Maybank, S.J., Tao, D.: Wide-angle image rectification: A
  survey. International Journal of Computer Vision  \textbf{130}(3),  747–776
  (3 2022). \doi{10.1007/s11263-021-01562-9}

\bibitem{fish-eye-entrance-pupil}
Fasogbon, P., Aksu, E.: Calibration of fisheye camera using entrance pupil (09
  2019). \doi{10.1109/ICIP.2019.8803832}

\bibitem{division-model}
Fitzgibbon, A.W.: Simultaneous linear estimation of multiple view geometry and
  lens distortion. In: Computer Society Conference on Computer Vision and
  Pattern Recognition. vol.~1 (2001). \doi{10.1109/CVPR.2001.990465}

\bibitem{lifcal}
Fleith, A., Ahmed, D., Cremers, D., Zeller, N.: Lifcal: Online light field
  camera calibration via bundle adjustment. In: German Conference on Pattern
  Recognition (8 2024)

\bibitem{robust-to-incomplete-pattern}
Gao, Z., Zhu, M., Yu, J.: A novel camera calibration pattern robust to
  incomplete pattern projection. IEEE Sensors Journal  \textbf{21}(8),
  10051--10060 (2021). \doi{10.1109/JSEN.2021.3058747}

\bibitem{speckle-pattern-on-display}
Genovese, K.: Single-image camera calibration with model-free distortion
  correction. Optics and Lasers in Engineering  \textbf{181} (10 2024).
  \doi{10.1016/j.optlaseng.2024.108348}

\bibitem{general-imaging-model}
Grossberg, M.D., Nayar, S.K.: A general imaging model and a method for finding
  its parameters. In: International Conference on Computer Vision. vol.~2, pp.
  108--115 (2001). \doi{10.1109/ICCV.2001.937611}

\bibitem{raxel-imaging-model}
Grossberg, M.D., Nayar, S.K.: The raxel imaging model and ray-based
  calibration. International Journal of Computer Vision  \textbf{61},  119--137
  (2 2005). \doi{10.1023/B:VISI.0000043754.56350.10}

\bibitem{deltille-grids}
Ha, H., Perdoch, M., Alismail, H., Kweon, I.S., Sheikh, Y.: Deltille grids for
  geometric camera calibration. In: International Conference on Computer
  Vision. pp. 5354--5362 (2017). \doi{10.1109/ICCV.2017.571}

\bibitem{distortion-polynomial-inequalities}
Heller, J., Henrion, D., Pajdla, T.: Stable radial distortion calibration by
  polynomial matrix inequalities programming. In: Computer Vision -- ACCV 2014.
  pp. 307--321 (1 2015)

\bibitem{rmse}
Hodson, T.O.: Root-mean-square error (rmse) or mean absolute error (mae): when
  to use them or not. Geoscientific Model Development  \textbf{15}(14),
  5481--5487 (2022). \doi{10.5194/gmd-15-5481-2022}

\bibitem{review-wide-angle-camera}
Huai, J., Shao, Y., Jozkow, G., Wang, B., Chen, D., He, Y., Yilmaz, A.:
  Geometric wide-angle camera calibration: A review and comparative study. IEEE
  Sensors Journal  \textbf{24} (10 2024). \doi{10.3390/s24206595}

\bibitem{bootstrapping}
James, G., Witten, D., Hastie, T., Tibshirani, R.: An Introduction to
  Statistical Learning. Springer New York, 2 edn. (2021)

\bibitem{concentric-circles-detection}
Jiang, G., Quan, L.: Detection of concentric circles for camera calibration.
  In: International Conference on Computer Vision. vol.~1, pp. 333--340 (2005).
  \doi{10.1109/ICCV.2005.73}

\bibitem{kannala-brandt}
Kannala, J., Brandt, S.S.: A generic camera model and calibration method for
  conventional, wide-angle, and fish-eye lenses. IEEE Transactions on Pattern
  Analysis and Machine Intelligence  \textbf{28}(8),  1335--1340 (2006)

\bibitem{need-accurate-pattern}
Lavest, J.M., Viala, M., Dhome, M.: Do we really need an accurate calibration
  pattern to achieve a reliable camera calibration? In: Proceedings of the 5th
  European Conference on Computer Vision. vol.~1, p. 158–174 (1998)

\bibitem{levenberg}
Levenberg, K.: A method for the solution of certain non–linear problems in
  least squares. Quarterly of Applied Mathematics  \textbf{2},  164--168
  (1944). \doi{10.1090/QAM/10666}

\bibitem{using-flat-monitors}
Lu, M.T., Chuang, J.H.: Fully automatic camera calibration for principal point
  using flat monitors. In: International Conference on Image Processing. pp.
  3154--3158 (2018). \doi{10.1109/ICIP.2018.8451222}

\bibitem{luhmann-book}
Luhmann, T.: Nahbereichsphotogrammetrie. Wichmann Verlag (8 2023)

\bibitem{luhmann-accuracy}
Luhmann, T., Fraser, C., Maas, H.G.: Sensor modelling and camera calibration
  for close-range photogrammetry. Photogrammetry and Remote Sensing
  \textbf{115},  37--46 (2016). \doi{10.1016/j.isprsjprs.2015.10.006}

\bibitem{familiy-distortion-models}
Ma, L., Chen, Y., Moore, K.L.: A family of simplified geometric distortion
  models for camera calibration. ArXiv  \textbf{cs.CV/0308003} (2003).
  \doi{10.48550/arXiv.cs/0308003}

\bibitem{analytical-undistortion}
Ma, L., Chen, Y., Moore, K.L.: Flexible camera calibration using a new
  analytical radial undistortion formula with application to mobile robot
  localization (2003). \doi{10.48550/arXiv.cs/0307045}

\bibitem{marquardt}
Marquardt, D.W.: An algorithm for least-squares estimation of nonlinear
  parameters. Journal of the Society for Industrial and Applied Mathematics
  \textbf{11}(2),  431--441 (1963). \doi{10.1137/0111030}

\bibitem{smooth-models}
Miraldo, P., Araujo, H.: Calibration of smooth camera models. IEEE Transactions
  on Pattern Analysis and Machine Intelligence  \textbf{35},  2091--2103 (09
  2013). \doi{10.1109/TPAMI.2012.258}

\bibitem{parametric-general-model}
Miraldo, P., Araujo, H., Queiró, J.: Point-based calibration using a
  parametric representation of the general imaging model. In: International
  Conference on Computer Vision. pp. 2304--2311 (2011).
  \doi{10.1109/ICCV.2011.6126511}

\bibitem{apriltag}
Olson, E.: {AprilTag}: A robust and flexible visual fiducial system. In:
  Proceedings of the {IEEE} International Conference on Robotics and Automation
  ({ICRA}). pp. 3400--3407 (May 2011)

\bibitem{otsu-threshold}
Otsu, N.: A threshold selection method from gray-level histograms. IEEE
  Transactions on Systems, Man, and Cybernetics  \textbf{9}(1),  62--66 (1979).
  \doi{10.1109/TSMC.1979.4310076}

\bibitem{camera-pose-implicit-distortion}
Pan, L., Pollefeys, M., Larsson, V.: Camera pose estimation using implicit
  distortion models. In: Conference on Computer Vision and Pattern Recognition.
  pp. 12809--12818 (06 2022). \doi{10.1109/CVPR52688.2022.01248}

\bibitem{unifying-model}
Ramalingam, S., Sturm, P.: A unifying model for camera calibration. IEEE
  Transactions on Pattern Analysis and Machine Intelligence  \textbf{39}(7),
  1309--1319 (7 2017). \doi{10.1109/TPAMI.2016.2592904}

\bibitem{towards-generic-camera}
Ramalingam, S., Sturm, P., Lodha, S.K.: Towards complete generic camera
  calibration. In: Computer Society Conference on Computer Vision and Pattern
  Recognition. vol.~1, pp. 1093--1098 (2005). \doi{10.1109/CVPR.2005.347}

\bibitem{SantosPedroFraunhoferIGD2020A3MD}
Santos, P., Tausch, R., Domajnko, M., Ritz, M., Knuth, M., Fellner, D.W.:
  {A}utomated 3{D} {M}ass {D}igitization for the {GLAM} {S}ector. Archiving
  2020 online: IS\&T. - 2161-8798 (ISSN) 2168-3204 (E-ISSN). - (2020)
  (Archiving Conference), IS\&T (2020)

\bibitem{10000-parameters}
Schöps, T., Larsson, V., Pollefeys, M., Sattler, T.: Why having 10,000
  parameters in your camera model is better than twelve. In: Conference on
  Computer Vision and Pattern Recognition. pp. 2532--2541 (6 2020).
  \doi{10.1109/CVPR42600.2020.00261}

\bibitem{feature-extraction-reimagined}
Shi, Z.: Rotating-star pattern for camera calibration (5 2025).
  \doi{10.48550/arXiv.2410.13371}

\bibitem{projective-geometry-book}
Springer, C.E.: Geometry and Analysis of Projective Spaces. W. H. Freeman and
  Company (1 1964)

\bibitem{target-free-network}
Stamatopoulos, C., Fraser, C.S.: Automated target-free network orientation and
  camera calibration. Annals of the Photogrammetry, Remote Sensing and Spatial
  Information Sciences  \textbf{II-5},  339--346 (05 2014).
  \doi{10.5194/isprsannals-II-5-339-2014}

\bibitem{plane-based-calibration}
Sturm, P., Maybank, S.: On plane-based camera calibration: A general algorithm,
  singularities, applications. In: Computer Society Conference on Computer
  Vision and Pattern Recognition. vol.~1, pp. 432--437 (1999).
  \doi{10.1109/CVPR.1999.786974}

\bibitem{generic-concept-calibration}
Sturm, P., Ramalingam, S.: A generic concept for camera calibration. In:
  Lecture Notes in Computer Science. vol.~3022, pp. 1--13 (5 2004).
  \doi{10.1007/978-3-540-24671-8\_1}

\bibitem{review-scheimpflug}
Sun, C., Liu, H., Jia, M., Chen, S.: Review of calibration methods for
  scheimpflug camera. IEEE Sensors Journal  \textbf{2018},  1--15 (03 2018).
  \doi{10.1155/2018/3901431}

\bibitem{segmentation-model}
Sun, Q., Hou, Y., Tan, Q.: A new method of camera calibration based on the
  segmentation model. Optik  \textbf{124}(24),  6991--6995 (12 2013).
  \doi{10.1016/j.ijleo.2013.05.179}

\bibitem{border-following}
Suzuki, S., Abe, K.: Topological structural analysis of digitized binary images
  by border following. Computer Vision, Graphics, and Image Processing
  \textbf{30}(1),  32--46 (1985). \doi{10.1016/0734-189X(85)90016-7}

\bibitem{swinbank2006fibonacci}
Swinbank, R., James~Purser, R.: Fibonacci grids: A novel approach to global
  modelling. Quarterly Journal of the Royal Meteorological Society: A journal
  of the atmospheric sciences, applied meteorology and physical oceanography
  \textbf{132}(619),  1769--1793 (2006)

\bibitem{unified-distortion}
Tamaki, T., Yamamura, T., Ohnishi, N.: Unified approach to image distortion.
  In: International Conference on Pattern Recognition. vol.~2, pp. 584--587 (2
  2002). \doi{10.1109/ICPR.2002.1048370}

\bibitem{bundle-adjustment}
Triggs, B., McLauchlan, P.F., Hartley, R.I., Fitzgibbon, A.W.: Bundle
  adjustment -- a modern synthesis. In: Vision Algorithms: Theory and Practice.
  pp. 298--372 (2000)

\bibitem{collinear-points-pole-polar}
Wei, L., Zhang, G., Huo, J., Xue, M.: Novel camera calibration method based on
  invariance of collinear points and pole–polar constraint. Journal of
  Systems Engineering and Electronics  \textbf{34}(3),  744--753 (2023).
  \doi{10.23919/JSEE.2023.000074}

\bibitem{wolfram-alpha-point-line-distance}
Weisstein, E.: Point-line distance -- 3-dimensional. {From MathWorld---A
  Wolfram Web Resource},
  \url{https://mathworld.wolfram.com/Point-LineDistance3-Dimensional.html},
  accessed on 14.08.2025

\bibitem{overview-fish-eye-distortion}
Xu, J., Han, D.W., Li, K., Li, J.J., Ma, Z.Y.: A comprehensive overview of
  fish-eye camera distortion correction methods (5 2024).
  \doi{10.48550/arXiv.2401.00442}

\bibitem{circle-projection-model}
Yang, L., Zhou, F., Zhang, W., Liu, Y.: A novel camera calibration method based
  on circle projection model. Measurement  \textbf{222} (2023).
  \doi{10.1016/j.measurement.2023.113651}

\bibitem{focused-plenoptic-camera}
Zhang, C., Ji, Z., Wang, Q.: Unconstrained two-parallel-plane model for focused
  plenoptic cameras calibration (8 2016). \doi{10.48550/arXiv.1608.04509}

\bibitem{rapid-calibration-by-one-image}
Zhang, J., Yu, H., Deng, H., Chai, Z., Ma, M., Zhong, X.: A robust and rapid
  camera calibration method by one captured image. IEEE Transactions on
  Instrumentation and Measurement  \textbf{68}(10),  4112--4121 (2019).
  \doi{10.1109/TIM.2018.2884583}

\bibitem{zhang-calibration}
Zhang, Z.: A flexible new technique for camera calibration. IEEE Transactions
  on Pattern Analysis and Machine Intelligence  \textbf{22},  1330--1334 (12
  2000). \doi{10.1109/34.888718}

\end{thebibliography}

\end{document}